%% file: Report-1-Olsztyn.tex
\documentclass[]{book}
\usepackage{fancyhdr}
\usepackage{fancybox}
\usepackage{amsmath}
\usepackage{graphicx}
\usepackage{url}
\usepackage{amsbsy,amssymb}
\usepackage{amsmath}
\usepackage{hyperref}
\usepackage{color}
\usepackage{authblk}
\usepackage{fancyvrb}
\usepackage[caption = false]{subfig}
\usepackage{lscape}
\usepackage{etoolbox}
\usepackage{soul}
\usepackage{geometry}

\patchcmd{\thebibliography}{\chapter*}{\section*}{}{}

\newcommand{\qed}{\ifmmode$\Box$\else{\unskip\nobreak\hfil
\penalty50\hskip1em\null\nobreak\hfil$\Box$
\parfillskip=0pt
\finalhyphendemerits=0\endgraf}\fi}
\newtheorem{Pa}{Paper}[section]

\newtheorem{Tm}[Pa]{{\bf Theorem}}
\newtheorem{Lm}[Pa]{{\bf Lemma}}

\newtheorem{Rm}[Pa]{{\bf Remark}}

\title{Selected aspects of complex, hypercomplex and fuzzy neural networks}
\author[1]{edited by Agnieszka Niemczynowicz}
\author[2,3]{Radosław A. Kycia}

\affil[1]{Faculty of Mathematics and Computer Science, University of Warmia and Mazury in Olsztyn, Poland}
\affil[2]{Faculty of Computer Science and Telecommunications, T. Kościuszko Cracow University of Technology, Kraków, Poland}
\affil[3]{Faculty of Science, Masaryk University, Brno, Czechia}

\date{January 3, 2023}

\begin{document}
\maketitle

\newpage
\tableofcontents

\newpage
\include{Intro}
\include{NKJ-1}

\include{NKJ-2}
\include{Siemaszko}

\include{Lluis}

\include{Baruch}

\include{Irina}

\include{7Artem}

\end{document}

%% file: Intro.tex
\chapter{Introduction}

This short report reviews the current state of the research and methodology on theoretical and practical aspects of Artificial Neural Networks (ANN). It was prepared to gather state-of-the-art knowledge needed to construct complex, hypercomplex and fuzzy neural networks.

The report reflects the individual interests of the authors and, by now means, cannot be treated as a comprehensive review of the ANN discipline. Considering the fast development of this field, it is currently impossible to do a detailed review of a considerable number of pages.

The report is an outcome of the Project Meeting\footnote{Project: 'The Strategic Research Partnership for the mathematical aspects of complex,  hypercomplex and fuzzy neural networks'} at the University of Warmia and Mazury in Olsztyn, Poland, organized in September 2022.

The contributors of the report are (in order of appearance):
\begin{itemize}
 \item{A. Niemczynowicz, UWM Olsztyn}
 \item{R.A. Kycia, CUT Cracow \& MUNI Brno}
 \item{M. Jaworski, CUT Cracow}
 \item{A. Siemaszko, UWM Olsztyn}
 \item{J.M. Calabuig, UPV Valencia}
 \item{Ll.M. Garc\'{\i}a-Raffi, UPV Valencia}
 \item{B. Schneider, UO Ostrava}
 \item{D. Berseghyan, UO Ostrava}
 \item{I. Perfiljeva, UO Ostrava}
 \item{V. Novak, UO Ostrava}
 \item{P. Artiemjew, UWM Olsztyn}
\end{itemize}

\section*{Acknowledgement}
The report has been supported by the Polish National Agency for Academic Exchange Strategic Partnership Programme under Grant No. BPI/PST/2021/1/00031.


%% file: NKJ-1.tex
\chapter{Biological inspiration of artificial neural networks}
\label{Chapter_1}
\addtocontents{toc}{\textit{R.A. Kycia, A. Niemczynowicz, M. Jaworski}\par}
\textit{R.A. Kycia, A. Niemczynowicz, M. Jaworski}

\section{Introduction}
\par The abilities of the human brain always inspired scientists to mimic its abilities. There are plenty of functionalities offered by this tissue. We can mention a few:
\begin{itemize}
\item Pattern recognition - recognition of objects, sounds, smells, touch.
\item Classification - distinguish similar objects.
\item Generation - digesting the input data and generating new output: movement, sound.
\end{itemize}
\par The structure of the brain at microscopic size was not revealed until the invention of the microscope and discovery of microorganisms by Van Leeuwenhoek in 1676. The distinction of neuron cells as a basic building block of neurons has also a long history that evolves with the ideas of how the neural tissue works. However the neuron theory, so called Neuron Doctrine, was proposed by Santiago Ram\'{o}n y Cajal (1852-1934). Since then the detailed study of biological and electrochemical properties of neurons were performed.
\par There are many different kinds of neurons depending on which part of the neural system we are analyzing. The complexity of the problem arises due to the complex structure of neurons and their mutual interactions. The mathematical description of the neuron in terms of, e.g., dynamical systems has a long history \cite{Izhikevich} and grows into the discipline of computational neurobiology \cite{Trappenberg}.
\par This also inspired computer scientists to use similar structures as neurons and nervous systems for computational tasks. In this section we present the history of these early attempts. The first attempt was the model of a single neuron as discussed below.
\section{Perceptron}
\par The mathematical ideas behind connection of neurons was presented in \cite{McCulloch}, and the first implementation of an artificial neuron, called perceptron, was designed at Cornell Aeronautical Laboratory by F. Rosenblatt as an electrical circuit and described in the report \cite{Rosenblatt_1}. However, in the appendix of the report the mathematical equations for the perceptron are provided.
\par In the simplest version the perceptron is able to distinguish two sets of data that can be separated by a linear hyperplane. In mathematical terms the input contains a $n$-dimensional vector of real numbers $X_{i}=[x_1, \ldots, x_n]$, and the associated to this feature that can be coded into a number $f_{i}$. The perceptron is characterized as a vector of $n$ weights (real numbers) $w=[w_1, \ldots, w_n]$ and an additional number $w_0$ called the bias unit. In order to make the computations more uniform the input vetor is extended to $X_{i}=[1, x_{1}, \ldots, x_{n}]$ and the weight vector to $W=[w_{0}, w_{1}, \ldots, w_{n}]$. The next element of the perceptron is a decision function, which can be the Heaviside step function $\theta(x)=1$ if $x>0$, $\theta(0)=0$ and $\theta(x)=-1$ if $x<0$. Then the data are classified with respect to the side of the hyperplane given by the equation $xw=0$, by $y=\theta(xw)$ function. The position of the plane (weights $w$) are taken such that to minimize the distance between the features $f_{i}$ and corresponding outputs $y_{i}=\theta(WX_{i})$. The implementation of the perceptron can be found in every standard machine learning books, e.g., \cite{Mirjalili}.
\par In the very core of the perceptron idea is that the effective classification is possible if the data are linearly separable, i.e., separable by a hyperplane. This idea was proved in \cite{Minsky} and is related to the XOR problem. Since the XOR gate can be recognized as a function that takes as an input two bits and returns the one bit, the input data for the function can be treated as input data for perceptron, and the features can be output of XOR function. It is obvious that these data are not linearly separable, and therefore, there is no possibility that perceptrons separate features $0$ from $1$ be a hyperplane (in fact, a line on the plane). The ideas in the book were influential enough to initiate, so called, (first) \textit{AI winter}, the period around 1980s where the perceptron idea was put on hold in favor of other AI architectures.
\section{Multilayer neural networks and backpropagation algorithm}
\par The problem of not linearly separated data was proven to be solved by showing that the connection of a few perceptrons in a network called neural network, and learning by back-propagating the error from the output through the network updating the weights is a solution of XOR problem. These ideas were described in \cite{Rumelhart_1} and started a renovation of the interests in artificial neurons. However, due to insufficient computing power needed for backpropagation algorithm the practical works were stalled up to 2010s and this period is called second AI winter.
\par The appearance of connected artificial neurons gave rise to the new paradigm of computing called connectionism that computing can be included in the topology (graph of connections between neurons in a ANN), see \cite{Buckner}.
\section{Current state of development}
\par Since the 2010s the increased interest in deep learning, i.e., use of neural networks for practical computational tasks is reviving. This situation is due to an increasing number of examples where neural networks can handle data achieving better performance than classical algorithms. This progress is also induced by the high interests in deep learning from the biggest IT companies.
\par The most common architectures used in applications are multilayer neural networks, where neurons are grouped in layers and each layer’s output is an input to another layer.
\par The simplification of construction of multilayer networks was provided by the appearance of two powerful OpenSource libraries:
\begin{itemize}
\item TensorFlow (Google Brain Team \cite{Google_1}) - donated by Google; For common use it is accessed by Keras frontend \cite{Chollet}.
\item PyTorch (Meta AI \cite{Meta_1}) - donated by Adam Paszke, Sam Gross, Soumith Chintala, and Gregory Chanan.
\end{itemize}
These two libraries have enormous infrastructure that was built around them and are considered as an industrial standard of deep learning.
\par As for the theoretical side of description of the neural networks, the insight in the idea of work of NN is expressed in, so called, Universal Approximation Theorems. The first was proved by G. Cybenko in 1989 \cite{Cybenko} for sigmoidal activation function. It was soon realized that the approximation properties of neural networks rest in multilayer (feedforward) architecture \cite{Hornik1989}, \cite{Hornik1991}.  Roughly stated, Universal Approximation Theorem says that for a specific topology (e.g., for arbitrary width and bounded depth) the output functions of feed-forward neural networks are dense in the space of continuous functions on a compact space and with supremum norm. Recent research in this direction focuses on estimating the optimal width and depth of layers to obtain best approximation properties, see e.g., \cite{Park} for further references.
\par There are many unresolved issues related to the work of ANN that are related to the issue of which functions can be approximated by ANN or the properties of learning algorithms. They are summarized in the review article \cite{Weinan}. As it was pointed out, at the current state of understanding there are still many unknowns however the big picture starts to emerge.
\section{Summary}
\par Currently, the deep learning progress is motivated by numerous applications starting from computer vision, natural language processing, self-driving cars, and ending to generation of multimedia (pictures sounds), or design pharmaceutics. The development is pushed by a big industry that has access to great computing powers needed to run learning algorithms. However, the theoretical description, improvements of algorithms, or construction of nonstandard (e.g., not multilayers) architecture is still possible within the limiting computing resources. As of 2022 it is still an active and promising area of research.
\section*{Acknowledgements}
The article has been supported by the Polish National Agency for Academic Exchange Strategic Partnership Programme under Grant No. BPI/PST/2021/1/00031.

%

%% file: NKJ-2.tex
\chapter{Classical architecture of artificial neural networks}
\addtocontents{toc}{\textit{A. Niemczynowicz, R.A. Kycia, M. Jaworski}\par}
\textit{A. Niemczynowicz, R.A. Kycia, M. Jaworski}

\vspace{5mm}

\section{Introduction}
\par Artificial Neural Networks (ANN) and Deep Learning discipline are currently the one of the most active fields of research in computer science. This activity is also inspired by a large demand of IT business for new solutions and architectures that are suitable for solving new problems or solving more efficiently problems that are currently solvable by classical algorithms.
\par Apart from various architectures, there is an issue of how to encode data to make them acceptable as an input to ANN.
\par The chapter is organized as follows: in the next section we acknowledge standard input data for ANN. Then we review typical structures of multilayers ANN, which are currently the most used architectures. Finally we make a list of nonstandard architectures.

\section{Encoding of data}
\par There are various types of data in the world. Many of them have standard ways of encoding to the form that is suitable for ANN processing. We provide an example, and by no means extinguishable list of data types:
\begin{itemize}
\item  Multidimensional numerical data that are transformable to vector or matrix forms. They are described in various Machine Learning books, e.g., \cite{Raschka_2}
\item Images - transformable to matrix representation. See, e.g., \cite{Tripathi}
\item Text data - transformable to vectors of words, e.g., BagOfWords, TFIDF vectors, transformers based on neural networks; see, e.g., \cite{Lane}
\item Graph data - represented as, e.g., incidence matrix. See, e.g., \cite{Cui}
\end{itemize}

\section{Multilayer ANN}
\par The typical architecture used in industry is a (multilayer) feedforward architecture that consists of layers of neurons where output of one layer is the input of another one. The acceleration in this direction was heavily induced by the appearance of two Open Source libraries that allow the construction of complex ANN architectures. These libraries are:
\begin{itemize}
\item TensorFlow (\cite{Google_2}) - donated by Google company.
\item PyTorch (\cite{Meta}) - donated by Adam Paszke, Sam Gross, Soumith Chintala, and Gregory Chanan.
\end{itemize}
The general nomenclature in such architectures is as follows:
\begin{itemize}
\item Input layer - layer that gets the input data.
\item Output layer - layer that returns the result of processing.
\item Hidden layers - all layers between Input and Output layers.
\end{itemize}
Within these frameworks are possible various multilayer architectures and processing capabilities that will be outlined below.
\par {\it Fully connected multilayer ANN} - this is a network where all neurons in a layer are connected with all neurons in neighborhood layers. Advantage of this architecture is that all neurons ‘see’ the whole output of the preceding layer. The main disadvantage of this simple architecture is that the number of connections in ANN grows enormously with the increase of the number of neurons.
\par {\it Convolutional ANN  (CNN)} - the convolution is a mathematical operation of connecting neighbor input data (words when processing text, pixels when processing images) to feed neurons  with more complete yet local information. This makes sense when the data can be treated as elements of topological space where there is some notion of closure that represents some real objects, e.g., a group of neighbor pixels can represent a dog or bird; the context of the sentence is represented by a sentence of words usually in close proximity. The convolution in general is a tool for grouping ‘close’ data together at the input, and moreover to provide some notion of group invariance. In typical applications the convolution is realized by the kernels that are discrete versions of translational-invariant functions. However, the general idea is related to equivariance with respect to group action \cite{Finzi} et al. In typical architecture of CNN the first few layers are convolutional layers that combine data and as a result reduce dimensionality of the data.
\par {\it Recurrent Neural Networks (RNN)} \cite{Rumelhart_2} - these networks can be described as a discrete dynamical system with feedback. The network is feeded by sequence of data and the output (of hidden layers) form the previous step. This kind of ``recursive processing'' allows the network to see correlations between data from different steps. Therefore such networks are ideal for text processing or time series predictions. The big drawback of this architecture is the complicated process of learning. Since the learning process is iterative the gradient used in backward propagation algorithm is computed many times and this can makes it extremely small or to blow up due to numerical manipulations. This is the so-called vanishing and exploding gradient problem. These are typical problems with gradient-based learning algorithms when the number of layers increases. Hopfield neural networks are a special kind of RNN.
\par {\it Long Short-Term Memmory ANN (LSTM)} \cite{Hochreiter} - this is the network where each neuron has its own software memory unit. The processing can be represented as a sequence of steps and the input form the previous steps is used to modify output values by means of the use of the memory. LSTM can be used in processing data where connection between different portions of data is important.
Encoder-Decoder architecture  \cite{Sutskever}- this rather more abstract architecture consists of two neural networks one for coding data and the second for encoding. The output is the output from the decoder. This neural network is designed for coding of sequences, e.g., translating from one language to the other one. Wehn processing large volumes of data (e.g. text) the decoder can lose the main purpose of processing, and therefore the attention mechanism was invented \cite{Bahdanau}.
\par {\it Generative Adversarial Network (GAN)} \cite{Goodfellow} - is also an architecture consisting of G (the generative model) and D (the discriminative model). They are learned in tandem where D estimates the probability that the output comes from the training data rather than from G.
\par This ends our non-inclusive overview of typical architectures used in typical industrial applications. In the next section we present some other architectures that are used on smaller scale or in the research on ANN.
\section{Other architectures}
We can distinguish:
\begin{itemize}
\item Hopfield neural networks \cite{Hopfield_2}, \cite{HopfieldPaper_2} - they are modeled on a physical system of spins on a lattice. Due to these similarities statistical physics methods can be widely applied for this architecture.
\item Boltzmann machines \cite{Sherrington_2} - is another spin-based approach to neural networks.
\item 'Algebraic Neural Networks' - under this title we collected the typical multilayer architectures, where computation is done using real numbers instead of real numbers, e.g., complex numbers, various Clifford algebras, and Hypercomplex algebras. This is currently a vast field of theoretical and practical research. An introduction to the research in this direction is presented in \cite{Arena}.
\end{itemize}
\section{Summary}
Due to many and still growing number of applications, ANN is an active and extremely promising research area. Therefore each summary is burdened with incompleteness and risk of fast outdating. In this chapter, the general overview of current architectures of ANN was presented with short characteristics.

\section*{Acknowledgement }

The article has been supported by the Polish National Agency for Academic Exchange Strategic Partnership Programme under Grant No. BPI/PST/2021/1/00031.

%% file: Siemaszko.tex
\chapter{Dynamical systems approach to artificial neural networks}
\addtocontents{toc}{\textit{R.A. Kycia, A. Siemaszko}\par}
\textit{R.A. Kycia, A. Siemaszko}

\section{Introduction}
Data processing in Neural Networks (biological and artificial) can described as a time-dependent phenomenon. The mathematical tool to describe the change of a system in time is offered by Dynamical Systems: Smooth Dynamical systems -- for a continuous time parameter usually ranging from a connected subset of $\mathbb{R}$, or Discrete Dynamical Systems -- discrete time steps, varying over a subset of $\mathbb{Z}$. Therefore, it is natural to ask if these complex systems can be described by the tools offered by Dynamical Systems. Currently, there are many directions in which the disciplines of Dynamical Systems and Artificial Neural Networks interpenetrate each other, and in this report, we indicate some of these directions in this fast-pacing field.

\section{Biological Neural Networks}
To understand the motivation for applying the Dynamical Systems approach to Artificial Neural Networks (ANN), we will briefly overview the modeling of biological neural networks. This is a vast subject of Dynamical Neuroscience ('neurodynamics'), see, e.g.,  \cite{DynamicalSystemsBiologicalNeurons}, or Chapter 21 of \cite{DynamicaSystemsPython}.

Even a single biological neuron is a very complicated electro-biochemical system. The main focus is on modeling neuron excitations -- when an electrochemical impulse passes some threshold, then the neuron 'fires', producing a sequence of spikes in voltage transmitted to other neurons by interconnectors called synapses. This modeling must take into account the self-sustaining states of inactivity and this producing spikes. In terms of Dynamical Systems, they can be modeled by limit cycles (attracting or repelling). The standard model for describing these phenomena is a Hodgkin-Huxley model, a four-dimensional model for cell membrane voltage, sodium and potassium densities in a cell, and so-called leakage gating \cite{Hodgkin-Huxley}.

The phenomenon of oscillation between inactive and spiking states inspired some researchers to base the computation on such oscillatory behavior, e.g., \cite{Oscillatory_computers, Oscillatory_computers2}. Moreover, the threshold behavior was adapted in the first model of a neuron -- the perceptron \cite{Perceptron_S}.

\section{Physics-motivated Neural Networks}
One of the systems that can be significantly investigated using the qualitative and quantitative methods of Dynamical Systems is the Hopfield Neural Network \cite{Hopfield_S}. The system is modeled on the crystal lattice of spins, and powerful techniques of statistical physics are accessible for solving its parameters. They allow us to estimate the memory capacity and stability of memorized patterns. For example, for the continuous version of the Hopfield Model, the stability can be analyzed using a suitable Lyapunov function, e.g., Chapter 20 of \cite{DynamicaSystemsPython}. This network model was developed into a core layer in a multilayer feed-forward ANN \cite{HopfieldPaper_S}.

The other physics-inspired model on spin glass is called the Boltzmann machine \cite{SpinGlass_S}. In this model also the techniques of statistical physics can be applied.

\section{Modelling Artificial Neural Networks}
Feed-forward multilayer ANN dominates current practical applications. The mathematical understanding of their work as a whole has yet to be provided, however, some progress is made \cite{MathematicalUnderstandingNN}.

The current trends of using Dynamical Systems theory to describe ANN focus on various directions, some of which we summarize in the following subsections.

\subsection{Modelling ANN work}
The description of ANN using continuous Dynamical Systems is a new idea \cite{Dynamics}. In principle, the multilayer ANN is a discrete Dynamical System, where we have two consecutive steps - performing a linear operation on the output from the previous step\footnote{The initial step is fed by the data.}, and applying a nonlinear function. We can now devise the idea to model such a network by a continuous Dynamical System. The original network is recovered by doing a discretization of the model. This approach is more flexible since powerful mathematical techniques are at our disposal. The problem of regression using the ODE approach can be formulated as follows \cite{Dynamics}: Consider the differential equation
\begin{equation}
 \frac{dz}{dt}=f(A(t), z), \quad z(0)=x,
\end{equation}
where $z$ and $f$ are $\mathbb{R}^{d}$-valued functions, $A$ is a control that need to be found. The solution of this problem $z(t)$, under linear transformation $u(x)=az(x)+b$, for real parameters $a \in\mathbb{R}^{d}$, $b\in\mathbb{R}$, must fit the data $y(x)$, i.e., to minimize the distance $||y(x)-u(x)||$ in a suitable norm. The information about the structure of ANN is contained in the function $f$. The problem of the existence of control at the level of Dynamical Systems is transferred in this approach to the question if the structure of ANN is suitable for modeling the data.

Especially interesting in terms of Dynamical Systems are Residual Neural Networks, which can be brought to discrete dynamical system \cite{Dynamics}, and in some cases, this system can be modeled as the Euler scheme for integrating ODEs \cite{ResNetDynamics}. Moreover, the Residual Model can be rewritten as a control problem of transport equation and then rewritten as a PDE on manifold \cite{ResNetPDEManifold}.

\subsection{Modelling of learning process}
The other aspect of the Dynamical System approach is the way how they model the learning algorithm. The ANN learning algorithm aims to find a minimum\footnote{In an ideal situation, it should be a global minimum.} for a loss function. This problem is usually solved by a gradient descent (GD) method. This method in hydrodynamical limit and using mean-field approximation \cite{MathematicalUnderstandingNN, GDFlow}, can be converted into a gradient flow of ANN weight on a manifold with the Wasserstein metric. This provides new mathematical tools for determining the convergence of DG methods.

In general, for the continuous approach to Learning ANN, one obtains nonlinear parabolic PDEs, where all tools from their theory, including optimal choice of function space, variational calculus, finding an approximate solution, analyzing the stability and attractors, can be applied, for reference see \cite{LearningContionus}.

Another approach to learning is the so-called Deep Equilibrium Model \cite{DeepEquilibriumModel}. In this approach, the learning is attained by finding the equilibrium of a Dynamical System that describes ANN.

\subsection{Neural ODEs}
Another approach in modeling ANN with ODEs are Neural ODEs, the concept presented in \cite{NeuralODEs}. The idea behind the model is to make the layers continuous. Then the propagation through the network can be described by ODE and not a difference equation. This opens an opportunity to apply adaptive ODE solvers for learning. The drawback of this approach is the limited approximation capabilities of these architectures, as described in \cite{NeuralODELearnignCapabilities}.

\section{Dynamical Models predicted by ANN}
The opposite direction of using ANN to model and control Dynamical Systems is currently a vast field of research. We do not pretend to review this field and only provide a reference of review book \cite{DSModellingByANN} instead.

\section{Conclusions}
Currently, the Artificial Neural Networks and Dynamical Systems theory merge, benefiting both disciplines. We presented some current trends in this direction, however, the full review is impossible due to the high volume of results appearing each term.

\section*{Acknowledgments}
The article has been supported by the Polish National Agency for Academic Exchange Strategic Partnership Programme under Grant No. BPI/PST/2021/1/00031.



%% file: Lluis.tex
\chapter{Neural networks as universal approximators}
\addtocontents{toc}{\textit{J.M. Calabuig, Ll.M. Garc\'{\i}a-Raffi}\par}
{\textit{J.M. Calabuig, Ll.M. Garc\'{\i}a-Raffi}}

\vspace{5mm}

Since the first golden age (the 1950s and 1960s) when in 1962, Frank Rosenblatt introduced and developed the perceptron, Artificial Neural Networks (ANNs) have gone through various stages ranging from enthusiasm to ostracism. When we talk about ANNs, we are talking about mathematical tools that play an important role in approximation and classification problems. From a mathematical point of view, a natural question that arises is whether Artificial Neural Networks are universal approximators in the sense of mathematics. This question, which may seem trivial or second-order in view of the applications of ANNs in applied problems, is nevertheless a central issue. In essence, the certainty of the results achieved in practical problem solved with Artificial Neural Networks rests on the certainty that they are universal approximators. To find the first answer to this question  we have to go back to the work of Cybenko and Hornik \cite{Cybenko1989,HORNIK1989359} where basically it is proved that a feed-forward Neural Networks with at least one hidden layer can approximate any continuous function  assuming that certain activation functions are used (sigmoid activation function). Since then, and as new network topologies emerged with new activation functions, an important theoretical effort have been done in order to prove the character of universal approximators of ANN \cite{LESHNO1993861,pinkus_1999,ZHOU2020787,Schafer,Heinecke,NEURIPS2020_e4acb4c8}.

Within the question of whether an ANN can approximate a (continuous) function there are two issues to be addressed. On the one hand, there is the Hornik/Cybenko issue which corresponds to the question about if some ANN can approximate a given (continuous) function to arbitrary precision. However, neither the result nor the proof of it give any indication of how \lq\lq large\rq\rq  ANNs need to be to achieve a certain approximation accuracy. Then, another issue to be addressed is  how many layers and how many neurons per layer an ANN requires, that is, the approximation rates. A distinction must be made between shallow learning and deep learning.

In \cite{SIEGEL2020313} authors study and proof approximation results for ANN with general activation functions: a two layer Neural Network with a polynomially-decaying non-sigmoidal activation function. They extend the results for a larger class of activation functions, removing the polynomial decay assumption. This result applies to any bounded, integrable activation function.

In \cite{pmlr-v75-yarotsky18a} authors address the study of the approximation of continuous functions with very deep networks using the activation function RELU. In this case, not narrow networks (a high number of neurons per layer) are considered
and authors prove that constant-width fully-connected networks of depth  of the order of the number of weights provide the fastest possible approximation rate.

In \cite{pmlr-v125-kidger20a} the narrow case is addressed, that is, networks of bounded width and arbitrary depth. Specially interesting is the work \cite{6252513} that address the super-narrow case, that is, with only two neurons per layer, showing that given enough layers, a super-narrow Neural Network, with two neurons per layer, is capable to separate any separable binary dataset and if the datasets exhibit certain type of symmetries, they are better suited for deep representation and may require only few hidden layers to produce desired classification.

Less literature is found on the consideration of non-standard activation functions. However, this is a field to be explored in order to obtain networks that are narrow, with a medium level of depth and a suitable approximation rate. Note that, for example, in traditional convolutional networks applied to the reconstruction of medical images (e.g. Nuclear Magnetic Resonance Imaging MRI), the number of weights (neurons+layers) is usually in the order of millions. In short, these are free parameters in our model and therefore any reduction in their number generates more robust and simpler mathematical models.

One of the natural extensions to changing the activation function is to consider that the image of the  function is not in $\mathbb{R}$ but in $\mathbb{C}$ \cite{NITTA19971391,KOBAYASHI2021535}. This is the case of complex and hypercomplex-valued Nerural Networks. Beyond being a simple generalization of real-value activation functions, Complex-Valued Neural Networks (CVNNs)  are specially suitable to deal with modelling problems of complex amplitude --amplitude and  phase-- the kind of problems that are in the core of wave physics (electromagnetism, light, sound/ultrasounds, and matter waves). CVNNs give an important advantage in practical applications in fields where signals  are massively analyzed and processed in time/space, frequency, and phase domains. Hyper-complex ANN as quaternion and Clifford Neural Networks are further extension of CVNNs (\cite{KOBAYASHI2017110, KOBAYASHI201738, Review2020, KOBAYASHI2021203, VIEIRA2022100032,DACUNHA2022126714}). They seems to be specially suitable in color-information treatment, image reconstruction and segmentation, robotics and systems control. The question about the character as universal approximates and the approximation rates of CVNNs  is currently the subject of investigation  \cite{Voigtlaender2020TheUA}, cf. \cite{arxiv.2209.02456}.

\vspace{5mm}
\section*{Acknowledgement }
The article has been supported by the Polish National Agency for Academic Exchange Strategic Partnership Programme under Grant No. BPI/PST/2021/1/00031.

%% file: Baruch.tex
\chapter{Complex and quaternionic neural networks}
\addtocontents{toc}{\textit{B. Schneider, D. Berseghyan}\par}
\textit{B. Schneider, D. Berseghyan}

\section{Introduction}
Neural networks in the real domain have been studied for a long time and achieved promising results in many vision tasks for recent years. However, the extensions of the neural network models in other number fields and their potential applications are not fully-investigated yet.

Complex numbers play an important role in practical applications and fundamental theorems in various fields of engineering such as electromagnetics, communication, control theory, and quantum mechanics. The application of complex numbers to neural networks has recently attracted attention because they tend to improve the learning ability and conform to the above mentioned applications.

They enable the modeling of a point in two-dimensional space as a single entity, rather than as a set of two data items on which $2$D geometrical affine operations are performed. It has been shown that a neural network with the representation and operations of complex numbers results in improved performance of the geometrical affine transformation in two-dimensional space, whereas the performance of real-valued (conventional) neural networks is comparatively poor. The operations involving complex numbers would improve the performance of neural networks for processing two-dimensional data, e.g. book \cite{1}.

In the 1870s, William Kingdon Clifford introduced his geometric algebra, building on earlier works of Sir William Rowan Hamilton and Hermann Gunther Grassmann. Clifford intended to describe the geometric properties of vectors, planes, and higher-dimensional objects. Most physicists encounter the algebra in the guise of Pauli and Dirac matrix algebras of quantum theory. Many roboticists or computer graphic engineers use quaternions for $3$D rotation estimation and interpolation, as it is too difficult for them to formulate homogeneous transformations of high-order geometric entities using a point-wise approach. They resort often to tensor calculus for multivariable calculus. Since robotics and engineering make use of the developments of mathematical physics, many beliefs are automatically inherited; for instance, some physicists come away from a study of Dirac theory with the view that Clifford’s algebra is inherently quantum-mechanical.

Extension of neural networks on hypercomplex number system is one of such research efforts. Input, output, and internal state of a neuron which is the basic computational unit are represented by hypercomplex number in these types of neural networks.

Quaternion neural networks are models in which computations of the neurons are based on quaternions, the four-dimensional equivalents of imaginary numbers. The quaternion neural network also performs superior in terms of convergence speed to a real-valued neural network with respect to the $3$-bit parity check problem, as simulations show. Consequently, the application of hypercomplex numbers, particularly quaternions, to neural networks has been investigated. Quaternions are a class of hypercomplex number systems, a four-dimensional extension of imaginary numbers. One of benefits by quaternions is that affine transformation of geometric figures in three-dimensional space ($3$D geometrical affine transformation), especially spatial rotation, can be represented compactly and efficiently; in recent years, quaternions are extensively used in the fields of robotics, control of satellites, and computer graphics, etc, see for example \cite{2}.

In that sense, it is thought that it is very useful to employ complex numbers and quaternions that can calculate two or three
dimensional information as a unit as expressions of neurons. In fact, it is suggested that complex-valued and quaternionic feed forward neural networks have a remarkable learning ability in terms of affine transformation problems in two or three dimensional space.  The role of Neural Networks in today’s scientific community cannot be denied, its vast applications, from engineering to medicine, these are based on continuously improving algorithms.

This motivated our research group to begin the approach towards the creation of a mathematical basis for the field of Hypercomplex Neural Networks which could bring better, faster algorithms, and be useful in a wide range of computations.

In the underlying mathematical theories, the choice of a system of constants plays an important role, and advancing from a theory built on real numbers to hypercomplex ones is bound to give improved algorithms, due to the rich analysis of the field.

\section{Elements of quaternionic analysis}

In this section we present briefly the basic definitions and results of quaternionic analysis which are necessary
for our purpose. For more information, we refer the reader to \cite{GS, Kravchenko}.

Let $\mathbb H$ be the set of real quaternions, i.e., that each quaternion $a$ is represented in the form $a= a_0 + a_1i + a_2j + a_3k$, with $\{a_k\}\subset\mathbb R, k = 0, 1, 2, 3$ and $i, j, k$ are the quaternionic imaginary units. The basic elements define arithmetic rules in $\mathbb H$, which are given by the following relations:
\[
i^2 = j^2= k^2 = -1;\: ij = k =-ji;\; jk = i = - kj;\; ki = j = -ik.
\]
The quaternionic conjugation of $a= a_0 + a_1i + a_2j + a_3k$ is given by $\bar a: = a_0 - a_1i - a_2j - a_3k$. It is easy seen that $a\bar a = \bar a a = a^2_0 + a^2_1 + a^2_2 + a^2_3$. Note that for $a,b\in\mathbb H,\,\overline{a b} = \bar b\bar a$.

We identify the space $\mathbb C^2$ with the set $\mathbb H$ of quaternions: let $(z_1, z_2) =
(x_0 + ix_1, x_2 + ix_3)$ be two complex numbers with the imaginary unit $i$, and let $j$ be another imaginary unit such that $j^2 = -1$ and $ij +ji = 0$ hold. In particular, for $a\in\mathbb C$ and by abuse of notation if $\bar{a}$ denoted the complex conjugate of $a$, we have $aj=j\bar{a}$. The set of elements of the form $q = z_1 + z_2j, z_1,z_2\in\mathbb C$, endowed both with a component-wise addition and with the associative multiplication is then another way of stating $\mathbb H$.  The quaternion conjugation gives: $\overline{z_1+z_2j} := {\bar{z}_1}-z_2j$ and $q\bar{q} = \bar{q}q = |z_1|^2 + |z_2|^2$.

Let $E$ be a bounded subset of $\mathbb R^4\cong\mathbb C^2\cong\mathbb C\times\mathbb C$ and denote by $BC(E,\mathbb H)$ the class of $\mathbb H$-valued bounded continuous functions on $E$. For $f\in BC(E,\mathbb H)$ we define the modulus of continuity of $f$ as a non-negative function $w_f(\delta), \delta>0$, given by
\[
w_f(\delta) := \sup_{|x-y|\leq \delta} \{|f(x)-f(y)|:\ x,y\in E\}.
\]
For $0<\nu\leq 1$ if
\[
\sup_{0<\delta\leq diam\ E}\left\{\frac{w_f(\delta)}{\delta^\nu}\right\} < \infty,\ \mbox{\rm for}\ \ 0<\nu\leq 1,
\]
then $f$ becomes a H\"older continuous with exponent $\nu$ function in $E$ (Lipschitz continuous for $\nu = 1$).
We will denote by
\[
C^{0,\nu}(E,\mathbb H) := \{f\in BC(E,\mathbb H):\ \ \sup_{0<\delta\leq diam\ E}\left\{\frac{w_f(\delta)}{\delta^\nu}\right\} < \infty\},
\]
the collection of H\"older continuous functions on $E$, for\ \ $0<\nu\leq 1$.

We say (\cite{David}) that a closed set $E$ in $\mathbb R^4$ is an Ahlfors-David regular set (in short AD-regular) if there exists a constant $c>0$ such that for all $x\in E$ and $0<r<\ diam\ E$ there holds
\[
c^{-1}r^3\leq \mathcal H^3(E\cap\mathbb B(x,r))\leq cr^3,
\]
where $\mathbb B(x,r)$ is closed ball with center $x$ and radius $r$ and $\mathcal H^3$ is the $3$-dimensional Hausdorff measure. The AD-regularity condition implies a uniform positive and finite bound on $E$ for the upper and lower density. Moreover, we notice that such condition produces a very wide class of surfaces that contains the classes of surfaces classically considered in the literature: Liapunov surfaces, smooth surfaces and Lipschitz ones. Finally we would like to remark that AD-regular sets are not always rectifiable in the sense of Federer \cite{Fe}.

In what follows, $\Omega\subset \mathbb R^4$ stands for a bounded domain with an AD-regular rectifiable boundary $\Gamma$ and let $\Omega^+:= \Omega$; $\Omega^- := \mathbb R^4\setminus\overline{\Omega^+}$, where both open sets are assumed to be connected.

For continuously real-differentiable function $\mathbb H$-valued functions $f:=f_0  + f_1 i+ f_2 j + f_3 k:\ \Omega\to\mathbb H$, the operator
\[
{}^\psi D := \frac{\partial }{\partial x_0}+i\frac{\partial }{\partial x_1}-j\frac{\partial }{\partial x_2}+k\frac{\partial }{\partial x_3},
\]
associated to the structural set $\mathbb H$-vector $\psi := \{1,i,-j,k\}$ is called the Cauchy–Riemann operator, which can be written in complex form as:
\[
{}^\psi D = 2\left\{\frac{\partial}{\partial {\bar z}_1} - j\frac{\partial}{\partial{\bar z}_2}\right\}
\]

A factorization of the Laplacian is given by
\[
{}^\psi D\, {}^{\bar\psi} D = {}^{\bar\psi} D\, {}^\psi D = \Delta_\mathbb H,
\]
where
\[
 {}^{\bar\psi} D := 2\left\{\frac{\partial}{\partial {\bar z}_1} + j\frac{\partial}{\partial{\bar z}_2}\right\},
\]
and $\Delta_\mathbb H [f] := \Delta_{\mathbb R^4} [f_0] + \Delta_{\mathbb R^4} [f_1] + \Delta_{\mathbb R^4} [f_2] + \Delta_{\mathbb R^4} [f_3]$.
A function $f:\ \Omega\to\mathbb H$ is called left $\psi-$hyperholomorphic in $\Omega$ if
\begin{eqnarray*}\label{CR1}
{}^\psi D[f](\xi) = 0\,\, \mbox{for}\,\, \forall\xi\in\Omega.
\end{eqnarray*}
We will denote by
\[
{}^\psi{\mathcal M}(\Omega,\mathbb H):=\{f\in C^1(\Omega,\mathbb H):\, {}^\psi D[f](\xi)=0,\,\forall\xi\in\Omega\}.
\]
Under assumption $f\in {}^\psi{\mathcal M}(\Omega,\mathbb H)$ and following similar arguments to those in \cite[page 3875]{ABAB} we have the Cauchy integral formula
\begin{eqnarray}\label{Cauchy1}
\int_\Gamma {\mathcal K}_\psi(\tau-t)\cdot n_\psi(\tau)\cdot f(\tau)\,d{\mathcal H}^3_\tau = f(t),\,t\in\Omega^+.
\end{eqnarray}
For a survey of the theory of $\psi$-hyperholomorphic functions along classical lines we refer the reader to \cite{Shapiro}.

An easy computation shows that if $f=u+vj$ with $u= f_0+if_1$ and $v=f_2+if_3$, then
\begin{equation*}
{}^{\psi}D f=0\,\Longleftrightarrow \,
\begin{cases}
\partial_{\bar z_1} u+\partial_{z_2}\bar{v} =0\cr
\partial_{\bar z_2} u-\partial_{z_1}\bar{v} =0,
\end{cases}
\end{equation*}
which express the direct relation between the $\psi$-hyperholomorphic functions and solutions of the Cimmino system.

The most important examples of a $\psi$-hyperholomorphic function is the function
\[
\mathcal K_{{\psi}}(q)=\frac{1}{2\pi^2}\frac{\bar{z_1}+\bar{z_2}j}{(|z_1|^2+|z_2|^2)^2},\, z_1, z_2\neq 0,
\]
which is obtained by applying ${}^{\bar\psi} D$ to the fundamental solution of the Lapacian $\Delta_{\mathbb R^4}$. It is known
as the Cauchy kernel and it represents a fundamental solution to both operators ${}^\psi D$ and ${}^{\bar\psi} D$.

\section{Poincar\'e-Bertrand formula for $\psi$-hyperholomorphic singular integrals}\label{section}
\setcounter{equation}{0}

The Cauchy kernel $\mathcal K_{{\psi}}$ generates important integrals for us:
\begin{itemize}
\item
\[
{}^{\psi}C_\Gamma[f](q) := \int\limits_\Gamma{\mathcal K_{{\psi}}}(\xi-q)\,n_{{\psi}}(\xi)\cdot f(\xi)\, d\mathcal H^3_\xi,\ q\in\mathbb R^4\setminus\Gamma,
\]
\item
\[
{}^{\psi}S_\Gamma[f](q) = 2\int\limits_\Gamma\mathcal K_{{\psi}}(\xi-q)\,n_{\psi}(\xi)\,(f(\xi)-f(q))\,d\mathcal H^3_\xi + f(q),\ q\in\Gamma,
\]
\end{itemize}
where $n_{\psi}(\xi) := n_0+n_1i-n_2j+n_3k$ with $(n_0,n_1,n_2,n_3)\in\mathbb R^4$ being the outward unit normal vector on $\Gamma$.

By using ideas from \cite{ABK}, we have
\begin{Rm}\label{rm1} In general, the integral
\[
\int_\Gamma \mathcal K_{{\psi}}(\xi-q)\,n_{\psi}(\xi)\,d\mathcal H^3_\xi
\]
has no sense for every $q\in\Gamma$, hence the formula
\[
\int_\Gamma \mathcal K_{{\psi}}(\xi-q)\,n_{\psi}(\xi)\,(f(\xi)-f(q))\,d\mathcal H^3_\xi =
\int_\Gamma \mathcal K_{{\psi}}(\xi-q)\,n_{\psi}(\xi)\,f(\xi)\,d\mathcal H^3_\xi - \left(\int_\Gamma \mathcal K_{{\psi}}(\xi-q)\,n_{\psi}(\xi)\,d\mathcal H^3_\xi\right)f(q)
\]
is generaly not valid. In case when the singular integral
\[
2\int_\Gamma \mathcal K_{{\psi}}(\xi-q)\,n_{\psi}(\xi)\,d\mathcal H^3_\xi
\]
has a finite value $\alpha(q)$ for $\forall q\in \Gamma$, then
\[
{}^{\psi}S_\Gamma[f](q) = 2\int\limits_\Gamma\mathcal K_{{\psi}}(\xi-q)\,n_{\psi}(\xi)\,f(\xi)\,d\mathcal H^3_\xi + (1-\alpha(q))f(q).
\]
\end{Rm}

While the first is a $\psi$-hyperholomorphic version of the usual Cauchy type integral the second represents its singular version, whose integral has to be taken in the sense of Cauchy's principal value.

In order to facilitate their usage, we present below some basic properties of the $\psi$-hyperholomorphic singular integrals, thus making our exposition self-contained.

\begin{Tm}\cite{ABB}\label{sp}
Let $\Omega$ be a bounded domain in $\mathbb R^4$ with AD-regular boundary $\Gamma$. Let $f\in C^{0,\nu}(\Gamma,\mathbb H)$. Then the following limits exist:
\begin{eqnarray*}
\lim_{\Omega^{\pm}\ni q\to \xi\in\Gamma}({}^{\psi}C_{\Gamma}[f](q))=:{}^{\psi}C^{\pm}_\Gamma[f](\xi),
\end{eqnarray*}
moreover the following identities hold:
\begin{eqnarray}\label{sp1}
{}^{\psi}C^{\pm}_\Gamma[f](\xi)=\frac{1}{2}[{}^{\psi}S_\Gamma[f](\xi)\pm f(\xi)],
\end{eqnarray}
for all $\xi\in\Gamma$.
\end{Tm}

\begin{Tm}\cite{ABB} \label{sq}
If $\Gamma$ is a AD-regular surface, then for $f\in C^{0,\nu}(\Gamma,\mathbb H), 0<\nu<1$ we have the following formula:
\begin{eqnarray*}
{}^{\psi}S^2_\Gamma[f](\xi) = f(\xi),\ \xi\in\Gamma.
\end{eqnarray*}
\end{Tm}

\begin{Lm}\label{lm1}
If $\{t,\tau\}\subset\Gamma,\, t\not=\xi$, then
\[
\int_{\Gamma_\tau}{\mathcal K}_{{\psi}}(\tau-t)\,n_{{\psi}}(\tau)\,{\mathcal K}_{{\psi}}(\tau-\xi)\, d{\mathcal H}_\tau^3 = 0.
\]
\end{Lm}
{\bf Proof.}\ The proof of Lemma \ref{lm1} is similar of the proof of Lemma 3 in \cite{Mitelman}, therefore we refer to \cite{Mitelman} for identical parts.

\begin{Lm}\label{lm4}
Let $f(\xi,\tau) := \frac{f_0(\xi,\tau)}{|\xi-\tau|^\mu},\ 0\leq\mu<3$, and $f_0\in C^{0,\nu}(\Gamma\times\Gamma,\mathbb H)$. Then the next formula of changing of integration order is true for all $t\in\Gamma$:
\[
\int_{\Gamma_\tau}{\mathcal K}_{{\psi}}(\tau-t)\,n_{{\psi}}(\tau)\,[f(\xi,\tau)-f(\tau,\tau)]\,d{\mathcal H}_\tau^3
\int_{\Gamma_\xi}n_{{\psi}}(\xi)\,d{\mathcal H}_\xi^3=
\]
\[
=\int_{\Gamma_\xi}\int_{\Gamma_\tau}{\mathcal K}_{{\psi}}(\tau-t)\,n_{{\psi}}(\tau)\,[f(\xi,\tau)-f(\tau,\tau)]\,d{\mathcal H}_\tau^3\,n_{{\psi}}(\xi)\,d{\mathcal H}_\xi^3.
\]
\end{Lm}
{\bf Proof.}\ The proof of Lemma\,\ref{lm4} is along the same line of the proof of Theorem 22.5 in \cite{Kytmanov}.\qed
The Poincar\'e-Bertrand formula in the $\psi$-hyperholomorphic framework is established by our next theorem.
\begin{Tm}\label{pb1}
Let $\Omega$ be a bounded domain in $\mathbb R^4$ with AD-regular boundary $\Gamma$ and let
$f\in C^{0,\nu}(\Gamma\times\Gamma,\mathbb H)$. Then for all $w\in\Gamma$
\[
\int_{\Gamma_z}{\mathcal K}_{{\psi}}(z-t)\,n_{{\psi}}(z)\,d{\mathcal H}_z^3
\int_{\Gamma_\xi}{\mathcal K}_{{\psi}}(\xi-z)\,n_{{\psi}}(\xi)[f(\xi,z)-f(z,t)]\,d{\mathcal H}_\xi^3=
\]
\[
= \int_{\Gamma_\xi}\int_{\Gamma_z}{\mathcal K}_{{\psi}}(z-t)\,n_{{\psi}}(z)\,d{\mathcal H}_z^3\,{\mathcal K}_{{\psi}}(\xi-z)\,n_{{\psi}}(\xi)[f(\xi,z)-f(z,t)]\,d{\mathcal H}_\xi^3 +\alpha^2(t)f(t,t),
\]
and the integrals being understood in the sense of the Cauchy principal value.
\end{Tm}
If $\Omega$ be a bounded domain in $\mathbb R^4$ with a smooth boundary  $\Gamma$ then $\alpha = \frac{1}{2}$ and the formula  reduces to the Poincar\'e-Bertrand formula (see, e.g., \cite{Mitelman}).

{\bf Proof.}\, Let
\[
\int_{\Gamma_\tau}{\mathcal K}_{{\psi}}(\tau-t)\,n_{{\psi}}(\tau)\,d{\mathcal H}_\tau^3
\int_{\Gamma_\xi}{\mathcal K}_{{\psi}}(\xi-\tau)\,n_{{\psi}}(\xi)\,[f(\xi,\tau)-f(\tau,t)]\,d{\mathcal H}_\xi^3 =
\]
\[
= \int_{\Gamma_\tau}{\mathcal K}_{{\psi}}(\tau-t)\,n_{{\psi}}(\tau)\,d{\mathcal H}_\tau^3
\int_{\Gamma_\xi}{\mathcal K}_{{\psi}}(\xi-\tau)\,n_{{\psi}}(\xi)\,([f(\xi,\tau)- f(\tau,t)] - f(\tau,\tau))\,d{\mathcal H}_\xi^3+
\]
\[
+ \int_{\Gamma_\tau}{\mathcal K}_{{\psi}}(\tau-t)\,n_{{\psi}}(\tau)\,d{\mathcal H}_\tau^3
\int_{\Gamma_\xi}{\mathcal K}_{{\psi}}(\xi-\tau)\,n_{{\psi}}(\xi)\,[f(\tau,\tau)-f(t,t)]\,d{\mathcal H}_\xi^3 +
\]
\[
+ \int_{\Gamma_\tau}{\mathcal K}_{{\psi}}(\tau-t)\,n_{{\psi}}(\tau)\,d{\mathcal H}_\tau^3
\int_{\Gamma_\xi}{\mathcal K}_{{\psi}}(\xi-\tau)\,n_{{\psi}}(\xi)\,f(t,t)\,d{\mathcal H}_\xi^3.
\]
In the first two quaternionic integrals on the right-hand side we can change the order of integration by Lemma \ref{lm4} we have
\[
\int_{\Gamma_\tau}{\mathcal K}_{{\psi}}(\tau-t)\,n_{{\psi}}(\tau)\,d{\mathcal H}_\tau^3\,
\int_{\Gamma_\xi}{\mathcal K}_{{\psi}}(\xi-\tau)\,n_{{\psi}}(\xi)\,[f(\xi,\tau)-f(\tau,t)] \,d{\mathcal H}_\xi^3 =
\]
\[
= \int_{\Gamma_\xi}\int_{\Gamma_\tau}{\mathcal K}_{{\psi}}(\tau-t)\,n_{{\psi}}(\tau)\,d{\mathcal H}_\tau^3\,
{\mathcal K}_{{\psi}}(\xi-\tau)\,n_{{\psi}}(\xi)\,([f(\xi,\tau)- f(\tau,t)] - f(\tau,\tau))\,d{\mathcal H}_\xi^3+
\]
\[
+ \int_{\Gamma_\xi}\int_{\Gamma_\tau}{\mathcal K}_{{\psi}}(\tau-t)\,n_{{\psi}}(\tau)\,d{\mathcal H}_\tau^3\,
{\mathcal K}_{{\psi}}(\xi-\tau)\,n_{{\psi}}(\xi)\,[f(\tau,\tau)-f(t,t)]\,d{\mathcal H}_\xi^3 +
\]
\[
+ \int_{\Gamma_\tau}{\mathcal K}_{{\psi}}(\tau-t)\,n_{{\psi}}(\tau)\,d{\mathcal H}_\tau^3\,
\int_{\Gamma_\xi}{\mathcal K}_{{\psi}}(\xi-\tau)\,n_{{\psi}}(\xi)\,f(t,t)\,d{\mathcal H}_\xi^3 =
\]
\[
= \int_{\Gamma_\xi}\int_{\Gamma_\tau}{\mathcal K}_{{\psi}}(\tau-t)\,n_{{\psi}}(\tau)\,d{\mathcal H}_\tau^3\,
{\mathcal K}_{{\psi}}(\xi-\tau)\,n_{{\psi}}(\xi)\,[f(\xi,\tau)-f(\tau,t)]\,d{\mathcal H}_\xi^3 -
\]
\[
- \int_{\Gamma_\xi}\,\left[\int_{\Gamma_\tau}{\mathcal K}_{{\psi}}(\tau-t)\,n_{{\psi}}(\tau)\,d{\mathcal H}_\tau^3\,
{\mathcal K}_{{\psi}}(\xi-\tau)\right]\,n_{{\psi}}(\xi)\,f(t,t)\,d{\mathcal H}_\xi^3 +
\]
\[
+ \int_{\Gamma_\tau}{\mathcal K}_{{\psi}}(\tau-t)\,n_{{\psi}}(\tau)\,d{\mathcal H}_\tau^3\,
\int_{\Gamma_\xi}{\mathcal K}_{{\psi}}(\xi-\tau)\,n_{{\psi}}(\xi)\,f(t,t)\,d{\mathcal H}_\xi^3 =
\]
(by using Lemma \ref{lm1} and the Remark \ref{rm1})
\[
= \int_{\Gamma_\xi}\int_{\Gamma_\tau}{\mathcal K}_{{\psi}}(\tau-t)\,n_{{\psi}}(\tau)\,d{\mathcal H}_\tau^3\,
{\mathcal K}_{{\psi}}(\xi-\tau)\,n_{{\psi}}(\xi)\,[f(\xi,\tau)-f(\tau,t)]\,d{\mathcal H}_\xi^3 +\alpha^2(t)f(t,t).
\]
\qed

Suppose that $f(\xi,\tau) = f(\xi)\in C^{0,\nu}(\Gamma,\mathbb H)$ is $\psi$-hyperholomorphic extension into $\Omega$, then the composition formula for $\psi$-hyperholomorphic functions can be written as:
\begin{Tm}(Composition formula)\,
Let $\Omega$ be a bounded domain in $\mathbb R^4$ with AD-regular boundary $\Gamma$ and let
$f\in C^{0,\nu}(\Gamma,\mathbb H)$. If $f(\xi)$ can be extended $\psi$-hyperholomorphically into $\Omega$. Then for all $t\in\Gamma$,
\begin{eqnarray}\label{copm}
\int_{\Gamma_\tau}{\mathcal K}_{{\psi}}(\tau-t)\,n_{{\psi}}(\tau)\,d{\mathcal H}_\tau^3
\int_{\Gamma_\xi}{\mathcal K}_{{\psi}}(\xi-\tau)\,n_{{\psi}}(\xi)[f(\xi)-f(\tau)]\,d{\mathcal H}_\xi^3 = \alpha^2(t)f(t).
\end{eqnarray}
\end{Tm}
Note that if
\[
{}^\psi\,{\tilde S} f:= 2\int_{\Gamma_\xi}{\mathcal K}_{{\psi}}(\xi-\tau)\,n_{{\psi}}(\xi)[f(\xi)-f(\tau)]\,d{\mathcal H}_\xi^3,
\]
than formula (\ref{copm}) means that
\[
{}^\psi\,{\tilde S}^2 f = 4\alpha^2(t) f(t).
\]
{\bf Proof.}\, Since $f(\xi)$ can be holomorphic extented into $\Omega$, then by Theorem \ref{sp} and Remark \ref{rm1}
\[
{}^\psi C^+f(z) = f(z),\,\, z\in\Omega.
\]
By formula (\ref{sp1}) and Remark \ref{rm1}, we have
\[
2f(z) = {}^\psi {\tilde S}f(\xi) + 2(1-\alpha(z))\,f(z),
\]
moreover
\[
{}^\psi {\tilde S}^2 f = {}^\psi {\tilde S}\,{}^\psi {\tilde S}f = {}^\psi {\tilde S}[2\alpha f] = 4\alpha^2 f.
\]
\qed
\section{Poincar\'e-Bertrand formula for the Cauchy-Cimmino singular integrals}\label{sec4}
\setcounter{equation}{0}

Using the representation of the quaternionic Cauchy kernel ${\mathcal K}_{{\psi}}$ and the normal vector $n_{\psi}$ in the complex form, we have:
\begin{eqnarray}\label{8}
{\mathcal K}_{{\psi}}(\xi-z)\,n_\psi(\xi) = K_1(\xi,z) + K_2(\xi,z)j,
\end{eqnarray}
with
\begin{eqnarray}\label{K1}
K_1(\xi,z) := \frac{1}{2\pi^2}\frac{({\bar \xi}_1-{\bar z}_1)(n_0+in_1) + ({\bar \xi}_2 - {\bar z}_2)(n_2+in_3)}{(|\xi_1-z_1|^2 + |\xi_2-z_2|^2)^2};
\end{eqnarray}
and
\begin{eqnarray}\label{K2}
K_2(\xi,z) := \frac{1}{2\pi^2}\frac{({\bar\xi}_2-{\bar z}_2){\overline{(n_0+in_1)}} - ({\bar\xi}_1-{\bar z}_1){\overline{(n_2+in_3)}}}{(|\xi_1-z_1|^2 + |\xi_2-z_2|^2)^2},
\end{eqnarray}
where $\xi =\xi_1 + \xi_2j,\, z = z_1+z_2j$.
Thus,
\begin{eqnarray*}
&&{}^{\psi}C_\Gamma[u+vj](z_1,z_2) = {\mathcal C}_{1}[u,v](z_1,z_2)+{\mathcal C}_{2}[u,v](z_1,z_2)j,\,\, (z_1,z_2)\notin\Gamma,\\
&&{}^{\psi}S_\Gamma[u+vj](z_1,z_2) ={\mathcal S}_{1}[u,v](z_1,z_2)+{\mathcal S}_{2}[u,v]j,\, \, (z_1,z_2)\in\Gamma,
\end{eqnarray*}
where
\[
\mathcal C_1[u,v](z_1,z_2)=
\int\limits_\Gamma\frac{[(\bar \zeta_{1}-\bar z_{1})(n_{0}+in_{1})+(\bar \zeta_{2}-\bar z_{2})(n_{2}+in_{3})]u(\zeta_1,\zeta_2)}
{2\pi^2(|\zeta_{1}-z_1|^2+|\zeta_{2}-z_2|^2)^2}\,d\mathcal H^3_{\xi_1,\xi_2} -
\]
\[
-\int\limits_\Gamma\frac{[(\bar \zeta_{2}-\bar z_{2})\overline{(n_{0}+in_{1})}-(\bar \zeta_{1}-\bar z_{1})\overline{(n_{2}+in_{3})}]\bar v(\zeta_1,\zeta_2)}{2\pi^2(|\zeta_{1}-z_1|^2+|\zeta_{2}-z_2|^2)^2}\,d\mathcal H^3_{\xi_1,\xi_2},
\]
\[
\mathcal C_2[u,v](z_1,z_2)
=\int\limits_\Gamma\frac{[(\bar \zeta_{1}-\bar z_{1})(n_{0}+in_{1})+(\bar \zeta_{2}-\bar z_{2})(n_{2}+in_{3})]v(\zeta_1,\zeta_2)
}{2\pi^2(|\zeta_{1}-z_1|^2+|\zeta_{2}-z_2|^2)^2}\,d\mathcal H^3_{\xi_1,\xi_2}+
\]
\[
+\int\limits_\Gamma\frac{[(\bar \zeta_{2}-\bar z_{2})\overline{(n_{0}+in_{1})}-(\bar \zeta_{1}-\bar z_{1})\overline{(n_{2}+in_{3})}]\bar u(\zeta_1,\zeta_2)}{2\pi^2(|\zeta_{1}-z_1|^2+|\zeta_{2}-z_2|^2)^2}\,d\mathcal H^3_{\xi_1,\xi_2}.
\]
The pair $(\mathcal C_1,\mathcal C_2)$ of integrals for $(z_1,z_2)\in\mathbb C^2$ play the role of an analog of a Cauchy type integral in theory of the Cimmino system of partial differential equations.

Similarly the singular Cauchy-Cimmino integral operators are defined formally as pair $(\mathcal S_1, \mathcal S_2)$, of the following singular integrals taken in the sense of Cauchy's principal value
\[
\mathcal S_1[u,v](z_1,z_2)=
2\int\limits_\Gamma\frac{[(\bar \zeta_{1}-\bar z_{1})(n_{0}+in_{1})+(\bar \zeta_{2}-\bar z_{2})(n_{2}+in_{3})][u(\zeta_1,\zeta_2)-u(z_1,z_2)]}{2\pi^2(|\zeta_{1}-z_1|^2+|\zeta_{2}-z_2|^2)^2}\,d\mathcal H^3-
\]
\[-2\int\limits_\Gamma\frac{[(\bar \zeta_{2}-\bar z_{2})\overline{(n_{0}+in_{1})}-(\bar \zeta_{1}-\bar z_{1})\overline{(n_{2}+in_{3})}][\bar v(\zeta_1,\zeta_2)-\bar v(z_1,z_2)]}{2\pi^2(|\zeta_{1}-z_1|^2+|\zeta_{2}-z_2|^2)^2}\,d\mathcal H^3+\]
\[
+u(z_1,z_2) =
\]
\[
= 2\int\limits_\Gamma\frac{[(\bar \zeta_{1}-\bar z_{1})(n_{0}+in_{1})+(\bar \zeta_{2}-\bar z_{2})(n_{2}+in_{3})]u(\zeta_1,\zeta_2)}{2\pi^2(|\zeta_{1}-z_1|^2+|\zeta_{2}-z_2|^2)^2}\,d\mathcal H^3-
\]
\[-2\int\limits_\Gamma\frac{[(\bar \zeta_{2}-\bar z_{2})\overline{(n_{0}+in_{1})}-(\bar \zeta_{1}-\bar z_{1})\overline{(n_{2}+in_{3})}]\bar v(\zeta_1,\zeta_2)}{2\pi^2(|\zeta_{1}-z_1|^2+|\zeta_{2}-z_2|^2)^2}\,d\mathcal H^3+\]
\[
+(1-\alpha(z_1,z_2))\,u(z_1,z_2),
\]
\[
\mathcal S_2[u,v](z_1,z_2)=2\int\limits_\Gamma\frac{[(\bar \zeta_{1}-\bar z_{1})(n_{0}+in_{1})+(\bar \zeta_{2}-\bar z_{2})(n_{2}+in_{3})][v(\zeta_1,\zeta_2)-v(z_1,z_2)]}{2\pi^2(|\zeta_{1}-z_1|^2+|\zeta_{2}-z_2|^2)^2}\,d\mathcal H^3+
\]
\[
+2\int\limits_\Gamma\frac{[(\bar \zeta_{2}-\bar z_{2})\overline{(n_{0}+in_{1})}-(\bar \zeta_{1}-\bar z_{1})\overline{(n_{2}+in_{3})}][\bar u(\zeta_1,\zeta_2)-\bar u(z_1,z_2)]}{2\pi^2(|\zeta_{1}-z_1|^2+|\zeta_{2}-z_2|^2)^2}\,d\mathcal H^3+\]
\[
+v(z_1,z_2) =
\]
\[
=2\int\limits_\Gamma\frac{[(\bar \zeta_{1}-\bar z_{1})(n_{0}+in_{1})+(\bar \zeta_{2}-\bar z_{2})(n_{2}+in_{3})]v(\zeta_1,\zeta_2)}{2\pi^2(|\zeta_{1}-z_1|^2+|\zeta_{2}-z_2|^2)^2}\,d\mathcal H^3+
\]
\[
+2\int\limits_\Gamma\frac{[(\bar \zeta_{2}-\bar z_{2})\overline{(n_{0}+in_{1})}-(\bar \zeta_{1}-\bar z_{1})\overline{(n_{2}+in_{3})}]\bar u(\zeta_1,\zeta_2)}{2\pi^2(|\zeta_{1}-z_1|^2+|\zeta_{2}-z_2|^2)^2}\,d\mathcal H^3+\]
\[
+
(1-\alpha(z_1,z_2))\,v(z_1,z_2).
\]
Returning to Section \ref{section} substitute (\ref{K1}) and (\ref{K2}) into Theorem \ref{pb1}, we have: let $\Omega$ be a bounded domain in $\mathbb R^4$ with AD-regular boundary $\Gamma$ and let
$(u,v)\in C^{0,\nu}(\Gamma\times\Gamma,\mathbb C)\times C^{0,\nu}(\Gamma\times\Gamma,\mathbb C)$, then for all $t\in\Gamma$
\[
\int_{\Gamma_\tau}\int_{\Gamma_\xi}
[K_1(\tau-t)+K_2(\tau-t)j]\,\left\{[K_1(\xi-\tau)+K_2(\xi-\tau)j](u(\xi,\tau)-u(\tau,t) + \right.
\]
\[
\left.+(v(\xi,\tau)-v(\tau,t))j)d\mathcal H^3_\xi\,d\mathcal H^3_\tau\right\} =
\]
\[
= \int_{\Gamma_\tau}\int_{\Gamma_\xi}
[K_1(\tau-t)+K_2(\tau-t)j]\,\left\{K_1(\xi-\tau)(u(\xi,\tau)-u(\tau,t)) + K_1(\xi-\tau)(v(\xi,\tau)-v(\tau,t))j  + \right.
\]
\[
\left.+K_2(\xi-\tau)j(u(\xi,\tau) - u(\tau,t)) + K_2(\xi-\tau)j(v(\xi,\tau)-v(\tau,t))j d\mathcal H^3_\xi\,d\mathcal H^3_\tau \right\} =
\]
\[
= \int_{\Gamma_\tau}\int_{\Gamma_\xi}\left\{K_1(\tau-t)\,K_1(\xi-\tau)(u(\xi,\tau) - u(\tau,t)) + \right.
\]
\[
+ K_1(\tau-t)\,K_1(\xi-\tau)(v(\xi,\tau) - v(\tau,t))j +
\]
\[
+ K_1(\tau-t)\,K_2(\xi-\tau)j(u(\xi,\tau) - u(\tau,t)) +
\]
\[
+ K_1(\tau-t)\,K_2(\xi-\tau)j(v(\xi,\tau) - v(\tau,t))j+
\]
\[
+ K_2(\tau-t)\,j\,K_1(\xi-\tau)(u(\xi,\tau)-u(\tau,t)) +
\]
\[
+ K_2(\tau-t)\,j\,K_1(\xi-\tau)(v(\xi,\tau) - v(\tau,t))j +
\]
\[
+ K_2(\tau-t)\,j\,K_2(\xi-\tau)\,j(u(\xi,\tau) - u(\tau,t)) +
\]
\[
+ \left.K_2(\tau-t)\,j\,K_2(\xi-\tau)\,j\,(v(\xi,\tau)-v(\tau,t))\,j\right\}d\mathcal H^3_\xi\,d\mathcal H^3_\tau.
\]
Note that
\[
\int_{\Gamma_\xi}\int_{\Gamma_\tau}{\mathcal K}_{{\psi}}(\tau-t)\,n_{{\psi}}(\tau)\,d{\mathcal H}_\tau^3\,{\mathcal K}_{{\psi}}(\xi-\tau)\,n_{{\psi}}(\xi)[f(\xi,\tau)-f(\tau,t)]\,d{\mathcal H}_\xi^3 +\alpha^2(t)f(t,t) =
\]
\[
= \int_{\Gamma_\xi}\int_{\Gamma_\tau}\left\{K_1(\tau-t)\,K_1(\xi-\tau)(u(\xi,\tau) - u(\tau,t)) + \right.
\]
\[
+ K_1(\tau-t)\,K_1(\xi-\tau)(v(\xi,\tau) - v(\tau,t))j +
\]
\[
+ K_1(\tau-t)\,K_2(\xi-\tau)j(u(\xi,\tau) - u(\tau,t)) +
\]
\[
+ K_1(\tau-t)\,K_2(\xi-\tau)j(v(\xi,\tau) - v(\tau,t))j+
\]
\[
+ K_2(\tau-t)\,j\,K_1(\xi-\tau)(u(\xi,\tau)-u(\tau,t)) +
\]
\[
+ K_2(\tau-t)\,j\,K_1(\xi-\tau)(v(\xi,\tau) - v(\tau,t))j +
\]
\[
+ K_2(\tau-t)\,j\,K_2(\xi-\tau)\,j(u(\xi,\tau) - u(\tau,t)) +
\]
\[
+ \left.K_2(\tau-t)\,j\,K_2(\xi-\tau)\,j\,(v(\xi,\tau)-v(\tau,t))\,j\right\}d\mathcal H^3_\tau\,d\mathcal H^3_\xi +
\]
\[
+\alpha^2(t)(u(t,t) + v(t,t)j).
\]
If one separates complex coordinates into above equality, then the following equalities can be easy obtained formulae for Cimmino system:
\[
\int_{\Gamma_\tau}\int_{\Gamma_\xi}[K_1(\tau-t)\,K_1(\xi-\tau)\,(u(\xi,\tau)-u(\tau,t)) +
\]
\[
+ K_1(\tau-t)\,K_2(\xi-\tau)\,j(v(\xi,\tau)-v(\tau,t))j +
\]
\[
+ K_2(\tau-t)\,j K_1(\xi-\tau)(v(\xi,\tau)-v(\tau,t))j +
\]
\[
+ K_2(\tau-t)\,j\,K_2(\xi-\tau)\,j(u(\xi,\tau)-u(\tau,t))]\,d{\mathcal H}^3_\xi\,d{\mathcal H}^3_\tau =
\]
\[
= \int_{\Gamma_\xi}\int_{\Gamma_\tau}[ K_1(\tau-t)\,K_1(\xi-\tau)\,(u(\xi,\tau) - u(\tau,t)) +
\]
\[
+ K_1(\tau-t)\,K_2(\xi-\tau)\,j(v(\xi,\tau)-v(\tau,t))j +
\]
\[
+ K_2(\tau-t)\,j\,K_1(\xi-\tau)(v(\xi,\tau)-v(\tau,t))j +
\]
\[
+ K_2(\tau-t)\,j\,K_2(\xi-\tau)\,j(u(\xi,\tau)-u(\tau,t))]\,d{\mathcal H}^3_\tau\,d{\mathcal H}^3_\xi +
\]
\[
+ \alpha^2(t)\,u(t,t);
\]
and
\[
\int_{\Gamma_\tau}\int_{\Gamma_\xi}[K_1(\tau-t)\,K_1(\xi-\tau)\,(v(\xi,\tau) - v(\tau,t))j +
\]
\[
+ K_1(\tau-t)\,K_2(\xi-\tau)\,j(u(\xi,\tau)-u(\tau,t)) +
\]
\[
+ K_2(\tau-t)\,j\,K_1(\xi-\tau)\,(u(\xi,\tau)-u(\tau,t)) +
\]
\[
+ K_2(\tau-t)\,j\,K_2(\xi-\tau)\,j(v(\xi,\tau)-v(\tau,t))j]\,d{\mathcal H}^3_\xi\,d{\mathcal H}^3_\tau =
\]
\[
= \int_{\Gamma_\xi}\int_{\Gamma_\tau}[K_1(\tau-t)\,K_1(\xi-\tau)\,(v(\xi,\tau)-v(\tau,t))j +
\]
\[
+ K_1(\tau-t)\,K_2(\xi-\tau)\,j(u(\xi,\tau) - u(\tau,t)) +
\]
\[
+ K_2(\tau-t)\,j\,K_1(\xi-\tau)\,(u(\xi,\tau) - u(\tau,t)) +
\]
\[
+ K_2(\tau-t)\,j\,K_2(\xi-\tau)\,j(v(\xi,\tau)-v(\tau,t))j]\,d{\mathcal H}^3_\tau\,d{\mathcal H}^3_\xi +
\]
\[
+\alpha^2(t)\,v(t,t)j.
\]
Here, we have
\begin{eqnarray}\label{ffpb1}
\int_{\Gamma_\tau}\int_{\Gamma_\xi}[K_1(\tau-t)\,K_1(\xi-\tau)\,(u(\xi,\tau) - u(\tau,t)) -
\end{eqnarray}
\[
- K_1(\tau-t)\,K_2(\xi-\tau)(\overline{v(\xi,\tau)-v(\tau,t)}) -
\]
\[
- K_2(\tau-t)\,\overline{K_1(\xi-\tau)}\,(\overline{v(\xi,\tau)-v(\tau,t)}) -
\]
\[
- K_2(\tau-t)\,\overline{K_2(\xi-\tau)}\,(u(\xi,\tau)-u(\tau,t))]\,d{\mathcal H}^3_\xi\,d{\mathcal H}^3_\tau =
\]
\[
=\int_{\Gamma_\xi}\int_{\Gamma_\tau}[K_1(\tau-t)\,K_1(\xi-\tau)\,(u(\xi,\tau) - u(\tau,t))-
\]
\[
- K_1(\tau-t)\,K_2(\xi-\tau)(\overline{v(\xi,\tau)-v(\tau,t)}) -
\]
\[
- K_2(\tau-t)\,\overline{K_1(\xi-\tau)}\,(\overline{v(\xi,\tau)-v(\tau,t)}) -
\]
\[
- K_2(\tau-t)\,\overline{K_2(\xi-\tau)}\,(u(\xi,\tau)-u(\tau,t))]\,d{\mathcal H}^3_\tau\,d{\mathcal H}^3_\xi +
\]
\[
+\alpha^2(t)u(t,t);
\]
and
\begin{eqnarray}\label{ffpb2}
\int_{\Gamma_\tau}\int_{\Gamma_\xi}[K_1(\tau-t)\,K_1(\xi-\tau)\,(v(\xi,\tau) - v(\tau,t)) +
\end{eqnarray}
\[
+ K_1(\tau-t)\,K_2(\xi-\tau)\,(\overline{u(\xi,\tau)-u(\tau,t)}) +
\]
\[
+ K_2(\tau-t)\,\overline{K_1(\xi-\tau)}\,(\overline{u(\xi,\tau)-u(\tau,t)}) -
\]
\[
- K_2(\tau-t)\,\overline{K_2(\xi-\tau)}\,(v(\xi,\tau)-v(\tau,t))]\,d{\mathcal H}^3_\xi\,d{\mathcal H}^3_\tau =
\]
\[
= \int_{\Gamma_\xi}\int_{\Gamma_\tau}[K_1(\tau-t)\,K_1(\xi-\tau)\,(v(\xi,\tau) - v(\tau,t)) +
\]
\[
+ K_1(\tau-t)\,K_2(\xi-\tau)\,(\overline{u(\xi,\tau)-u(\tau,t)}) +
\]
\[
+ K_2(\tau-t)\,\overline{K_1(\xi-\tau)}\,(\overline{u(\xi,\tau)-u(\tau,t)}) -
\]
\[
- K_2(\tau-t)\,\overline{K_2(\xi-\tau)}\,(v(\xi,\tau)-v(\tau,t))]\,d{\mathcal H}^3_\tau\,d{\mathcal H}^3_\xi +
\]
\[
+ \alpha^2(t)\,v(t,t).
\]

Let
\[
N_1[f](z) := 2\int_\Gamma K_1(\xi-z)\,f(\xi)\,d{\mathcal H}^3_\xi + (1-\alpha(z))f(z),\, \forall z\in\Gamma,
\]
and
\[
N_2[f](z) := -2\int_\Gamma K_2(\xi-z){\overline {f(\xi)}}\,d{\mathcal H}^3_\xi + (1-\alpha(z)){\overline{f(z)}},\,\forall z\in\Gamma.
\]
If one separates complex coordinates in Lemma \ref{lm1} we have:
\[
\int_{\Gamma_\xi}\int_{\Gamma_\tau} K_1(\tau-z)\,K_1(\xi-\tau)\,u(\xi)\,d{\mathcal H}^3_\tau\,d{\mathcal H}^3_\xi-
\int_{\Gamma_\xi}\int_{\Gamma_\tau} K_1(\tau-z)\,K_2(\xi-\tau)\,{\overline{v(\xi)}}\,d{\mathcal H}^3_\tau\,d{\mathcal H}^3_\xi-
\]
\[
-\int_{\Gamma_\xi}\int_{\Gamma_\tau} K_2(\tau-z)\,\overline{K_1(\xi-\tau)}\,\overline{v(\xi)}\,d{\mathcal H}^3_\tau\,d{\mathcal H}^3_\xi-
\int_{\Gamma_\xi}\int_{\Gamma_\tau} K_2(\tau-z)\,\overline{K_2(\xi-\tau)}\,u(\xi)\,d{\mathcal H}^3_\tau\,d{\mathcal H}^3_\xi = 0,
\]
\[
\int_{\Gamma_\xi}\int_{\Gamma_\tau} K_1(\tau-z)\,K_1(\xi-\tau)\,v(\xi)\,d{\mathcal H}^3_\tau\,d{\mathcal H}^3_\xi+
\int_{\Gamma_\xi}\int_{\Gamma_\tau} K_1(\tau-z)\,K_2(\xi-\tau)\,\overline{u(\xi)}\,d{\mathcal H}^3_\tau\,d{\mathcal H}^3_\xi +
\]
\[
+\int_{\Gamma_\xi}\int_{\Gamma_\tau} K_2(\tau-z)\,\overline{K_1(\xi-\tau)}\,\overline{u(\xi)}\,d{\mathcal H}^3_\tau\,d{\mathcal H}^3_\xi-
\int_{\Gamma_\xi}\int_{\Gamma_\tau} K_2(\tau-z)\,\overline{K_2(\xi-\tau)}\,v(\xi)\,d{\mathcal H}^3_\tau\,d{\mathcal H}^3_\xi = 0.
\]
Thus, if functions $u$ and $v$ depend only on $\xi$, then we can write:
\begin{eqnarray}\label{n1}
N^2_1 - N^2_2 = I,
\end{eqnarray}
\begin{eqnarray}\label{n2}
N_1\,N_2 + N_2\,N_1 = 0.
\end{eqnarray}
\begin{Rm}\label{r1}
Note that $N^2_2\not=0$ in (\ref{n1}). Indeed, if $N^2_2[f] =0$ for all $f\in C^{1,\nu}(\Gamma,\mathbb C)$. Then the function $N_2[f]$ can be holomorphically extended from $\Gamma$ into $\Omega^+$ and by the uniqueness theorem for harmonic functions this extension is given by
\[
F(z) = -2\int_\Gamma K_2(\xi-z)\overline{f(\xi)}\,d{\mathcal H}^3_\xi,\, z\in\Omega^+.
\]
But then ${}^\psi C[f]$ and (\ref{8}) imply that the function
\[
G_f(z) :=\int_\Gamma K_1(\xi-z)\,f(\xi)\,d{\mathcal H}^3_\xi,
\]
is holomorphic for any $f\in C^{1,\nu}(\Gamma,\mathbb C)$ which is not true.
\end{Rm}
\begin{Tm}
Let $f\in C^1(\overline{\Omega^+},\mathbb C)$ is representable in $\Omega^+\subset\mathbb C^2$ by
\[
f(z) = \int_\Gamma K_1(\xi-z)\,f(\xi)\,d{\mathcal H}^3_\xi,\,z\in\Omega^+.
\]
Then $f$ is holomorphic in $\Omega^+$.
\end{Tm}
{\bf Proof.}\ Applying the Sokhotski-Plemelj formulae to $f$ we have
\[
f(z) = \frac{1}{2}\,[N_1[f](z) + f(z)],\, z\in\Gamma.
\]
Thus, $N^2_1[f] = I[f]$, and from (\ref{n1}) we have $N^2_2[f] = 0$. But by Remark \ref{r1} we have that function $F$ defined by
\[
F(z) := \int_\Gamma K_1(\xi-z)\,f(\xi)\,d{\mathcal H}^3_\xi
\]
is holomorphic in $\Omega^+$ and with $F\mid\Omega^+ = u$ we completed the proof.\qed

From Lemma \ref{lm1} and \cite[page 211]{Kytmanov}, the term
\[
\int_{\Gamma_\xi}\int_{\Gamma_\tau} K_1(\tau-t)\,K_1(\xi-\tau)\,d\mathcal H^3_\tau\,d\mathcal H^3_\xi = 0\, \forall t\in\Gamma.
\]
Then from Section \ref{sec4}, we have that
\begin{eqnarray}\label{star1}
\int_{\Gamma_\xi}\int_{\Gamma_\tau} K_2(\tau-t)\,{\overline{K_2(\xi-\tau)}}\,d\mathcal H^3_\tau\,d\mathcal H^3_\xi = 0.
\end{eqnarray}
So, by using (\ref{star1}) and Theorem \ref{pb1}, for $t\in\Gamma$ and $f\in C^{0,\nu}(\Gamma\times\Gamma,\mathbb C)$ we have
\begin{eqnarray}\label{f19}
&&\int_{\Gamma_\tau}\int_{\Gamma_\xi} K_2(\tau-t)\,{\overline{K_2(\xi-\tau)}}\,[f(\xi,\tau)-f(\tau,t)]\,d\mathcal H^3_\tau\,d\mathcal H^3_\xi = \\
&&= \int_{\Gamma_\xi}\int_{\Gamma_\tau} K_2(\tau-t)\,{\overline{K_2(\xi-\tau)}}\,[f(\xi,\tau)-f(\tau,t)]\,d\mathcal H^3_\xi\,d\mathcal H^3_\tau.\nonumber
\end{eqnarray}
Comparing that last equality with (\ref{ffpb1}) and (\ref{ffpb2}), for $f\in C^{0,\nu}(\Gamma\times\Gamma,\mathbb C)$
for singular integrals for Cimmino system the structural analog of the Poin\'care-Bertrand formula is true:
\[
\int_{\Gamma_\tau}\int_{\Gamma_\xi}K_1(\tau-t)\,K_1(\xi-\tau)\,[f(\xi,\tau)-f(\tau,t)]d{\mathcal H}_\xi^3\,d{\mathcal H}_\tau^3  =
\]
\[
=\int_{\Gamma_\xi}\int_{\Gamma_\tau}K_1(\tau-t)\,K_1(\xi-\tau)\,[f(\xi,\tau)-f(\tau,t)]d{\mathcal H}_\tau^3\,d{\mathcal H}_\xi^3 + \alpha^2(t)f(t,t).
\]

\vspace{5mm}

\section*{Acknowledgement }

The article has been supported by the Polish National Agency for Academic Exchange Strategic Partnership Programme under Grant No. BPI/PST/2021/1/00031.

%% file: Irina.tex
\chapter{Fuzzy neural networks}
\addtocontents{toc}{\textit{I. Perfiljeva, V. Novak}\par}
\textit{I. Perfiljeva, V. Novak}

\vspace{5mm}

In the field of artificial intelligence, neuro-fuzzy refers to combination of artificial neural networks and fuzzy logic \cite{b4-1}.

A neuro-fuzzy system is commonly known in the literature as a fuzzy neural network (FNN) or a neuro-fuzzy system (NFS). A neuro-fuzzy system (used hereafter) incorporates the human reasoning style of fuzzy systems through the use of fuzzy sets and a linguistic model consisting of a set of fuzzy IF-THEN rules. The main strength of neuro-fuzzy systems is that they are universal approximators, the result of which allows interpretation by fuzzy IF-THEN rules \cite{b3,b4,b4-1}.

The main specificity of neuro-fuzzy systems is the presence of two conflicting requirements for fuzzy modeling: interpretability and accuracy. In practice, one of two requirements prevails.

As a consequence, the field of study of neuro-fuzzy systems is divided into two areas: linguistic fuzzy modeling focused on interpretability, mainly the Mamdani model; and accuracy-oriented fuzzy modeling, mainly the Takagi-Sugeno-Kangi (TSK) model.

A new line of research in the field of data flow mining considers the case when neuro-fuzzy systems are constantly updated with new incoming data. The system's response lies in its dynamic updates, including not only recursive adaptation of model parameters, but also dynamic evolution and reduction of model components to adequately handle concept drift and keep the model ``relevant'' at all times. Detailed reviews of various approaches to the development of neuro-fuzzy systems can be found in \cite{b3} and \cite{b4}.

A neuro-fuzzy system is represented as special three-layer feedforward neural network (ANN) where \cite{b4-1}
\begin{itemize}
\item The first layer corresponds to the input variables,
\item The second layer symbolizes the fuzzy rules,
\item The third layer represents the output variables,
\item The fuzzy sets are converted as (fuzzy) connection weights.
\end{itemize}
The learning procedure is constrained to ensure the semantic properties of the underlying fuzzy system.

Both characteristics: interpretability and accuracy become relevant when the NFS has already been successfully developed for solving a specific problem. This problem-oriented perspective exposes the limitations of NFS modeled by artificial neural networks (ANNs) and raises the question: can NFS be extended to the next generation of Convolutional Neural Networks (CNNs)? Below we consider the main problems specific to neural network computing technology.

The main problems solved with the help of neural networks are classification and regression. Other problems are their modifications. For example, semantic/instance segmentation is based on pixel-wise  classification; object detection is a regression on rectangle/polygon areas, time series prediction is a regression, etc.

Let us discuss and compare the capabilities of ANN and CNN in solving these problems \cite{b34,b44}. Both neural networks as computational models have a similar architecture with a common step --- feature extraction. The main difference is how they transform the input. From a CNN point of view, feature extraction is a gradual process focused on data with spatial dependencies. The convolution shifts its window over the data, which leads to invariance to data translation. Convolutions gradually extract many complex and abstract features. The result of this stage is a vector of descriptive features.

On the other hand, ANN feature extraction can be interpreted as a transformation of the input space into a space more suitable for a given task, i.e., making data samples separable. Consequently, ANN is typically used for data without spatial dependencies, such as tabular data.

The difference between ANN and CNN appears in different models of their computational units – neurons. ANN neuron output a is given as
\[
a=g(b+\sum_i w_i x_i)=g(b+ \mathbf{wx}),
\]

while the convolutional neuron output $a_ij$ is given as
\[
a_ij=g+b(\sum_{m=1}^l\sum_{n=1}^l \mathbf{W}_{m,n}
\mathbf{x}_{i+m,j+n}),
\]
where $\mathbf{W}$ is a convolutional kernel.

CNNs are currently the state-of-the-art models in all major computer vision tasks, from image classification and object detection to instance segmentation \cite{b22,b41,b44}. CNNs combine three architectural ideas: local receptive fields to extract elementary features from images; shared weights to extract the same set of elementary features from the entire input image and to lower computational costs; local averaging and sub-sampling to reduce the resolution of feature maps.

Typically, CNNs are built as a sequence of convolutional layers and pooled layers to automatically learn higher and higher level features \cite{b33,b44}.
At the end of the sequence, one or more fully connected layers are used to map the output feature map to the scores.

This structure entails complex internal relationships, which are difficult to explain using the Mamdani or Takagi-Sugeno type fuzzy models discussed above. Fortunately, the path to explainability for CNNs is easier than for other types of NN models, since human cognitive abilities contribute to the understanding of visual data. If we agree that the interpretability of a model is something that comes from the design of the model itself then \cite{b03}
\begin{quote}
\it an explainable AI is one that offers reasonable data processing details that make its operation clear or easy to understand.
\end{quote}

With this observation in mind, we single out one particular fuzzy modeling technique, known as fuzzy (F-)transforms, as a technique whose computational model is similar to the CNN model \cite{b15}.

It has been proven in many papers \cite{b11}--\cite{b15} that the higher degree F-transforms are universal approximators of smooth and discrete functions. The approximation on a whole domain is a combination  of locally best approximations called F-transform components. They are represented by higher degree polynomials and parametrized by coefficients that correspond to average values of local and nonlocal derivatives of various degrees. If the F-transform
is applied to images, then its parameters are used in regularization, edge detection, characterization of patches \cite{b17}, \cite{b37}, etc. Their computation can be performed by discrete convolutions with kernels that, up to the second degree, are similar to those widely
used in image processing, namely: Gaussian, Sobel, Laplacian \cite{b19}. Thus, we can draw an analogy with the CNN method of computation and call the parameters of the higher degree F-transform features. Moreover, based on a clear understanding of these features’ semantic meaning, we say that a CNN with the F-transform kernels extracts features with a clear interpretation. In addition, the sequential application of F-transform kernels with an up to the second degree gives average (nonlocal) derivatives of higher and higher degrees.

The following text details the neural network design supported by the theoretically proven F-transform methodology. The LeNet-5 architecture was chosen as the prototype architecture, and a new CNN called FTNet was compiled with kernels taken from the F-transform theory of the higher degree.

The performance of FTNet was examined on several datasets and on them it converges faster in terms of accuracy/loss than the baseline network, subject to the same number of steps. We compared the F-transform kernels in the first layer before and after training. We observed that the kernels remain unchanged. Moreover, their shapes are similar to the shapes of extracted kernel groups from the most known CNNs: VGG16 \cite{b41}, VGG19  \cite{b41},  InceptionV3 \cite{b44}, MobileNet \cite{b15}, ResNet  \cite{b15}, and AlexNet \cite{b22} as the representative examples of CNNs.

\section*{Acknowledgement }

The article has been supported by the Polish National Agency for Academic Exchange Strategic Partnership Programme under Grant No. BPI/PST/2021/1/00031.